\documentclass[runningheads]{llncs}
\usepackage[T1]{fontenc}
\usepackage{booktabs}
\usepackage{bm}
\usepackage{amsfonts}
\usepackage{graphicx,verbatim}
\begin{document}
\title{EndoPBR: Material and Lighting Estimation for Photorealistic Surgical Simulations via Physically-based Rendering}
%
\author{John J. Han\inst{1} \and
Jie Ying Wu\inst{1}}
\authorrunning{Han et al.}
%
\institute{Vanderbilt University, Nashville TN, 37212, USA\\
\email{john.j.han@vanderbilt.edu}}



\maketitle              
\begin{abstract}
The lack of labeled datasets in 3D vision for surgical scenes inhibits the development of robust 3D reconstruction algorithms in the medical domain. Despite the popularity of Neural Radiance Fields and 3D Gaussian Splatting in the general computer vision community, these systems have yet to find consistent success in surgical scenes due to challenges such as non-stationary lighting and non-Lambertian surfaces. As a result, the need for labeled surgical datasets continues to grow. In this work, we introduce a differentiable rendering framework for material and lighting estimation from endoscopic images and known geometry. Compared to previous approaches that model lighting and material jointly as radiance, we explicitly disentangle these scene properties for robust and photorealistic novel view synthesis. To disambiguate the training process, we formulate domain-specific properties inherent in surgical scenes. Specifically, we model the scene lighting as a simple spotlight and material properties as a bidirectional reflectance distribution function, parameterized by a neural network. By grounding color predictions in the rendering equation, we can generate photorealistic images at arbitrary camera poses. We evaluate our method with various sequences from the Colonoscopy 3D Video Dataset and show that our method produces competitive novel view synthesis results compared with other approaches. Furthermore, we demonstrate that synthetic data can be used to develop 3D vision algorithms by finetuning a depth estimation model with our rendered outputs. Overall, we see that the depth estimation performance is on par with fine-tuning with the original real images.\footnote{The codebase will be released at \url{https://github.com/juseonghan/EndoPBR}.}

\end{abstract}
\section{Introduction}

\begin{figure}
    \centering
    \includegraphics[width=\linewidth]{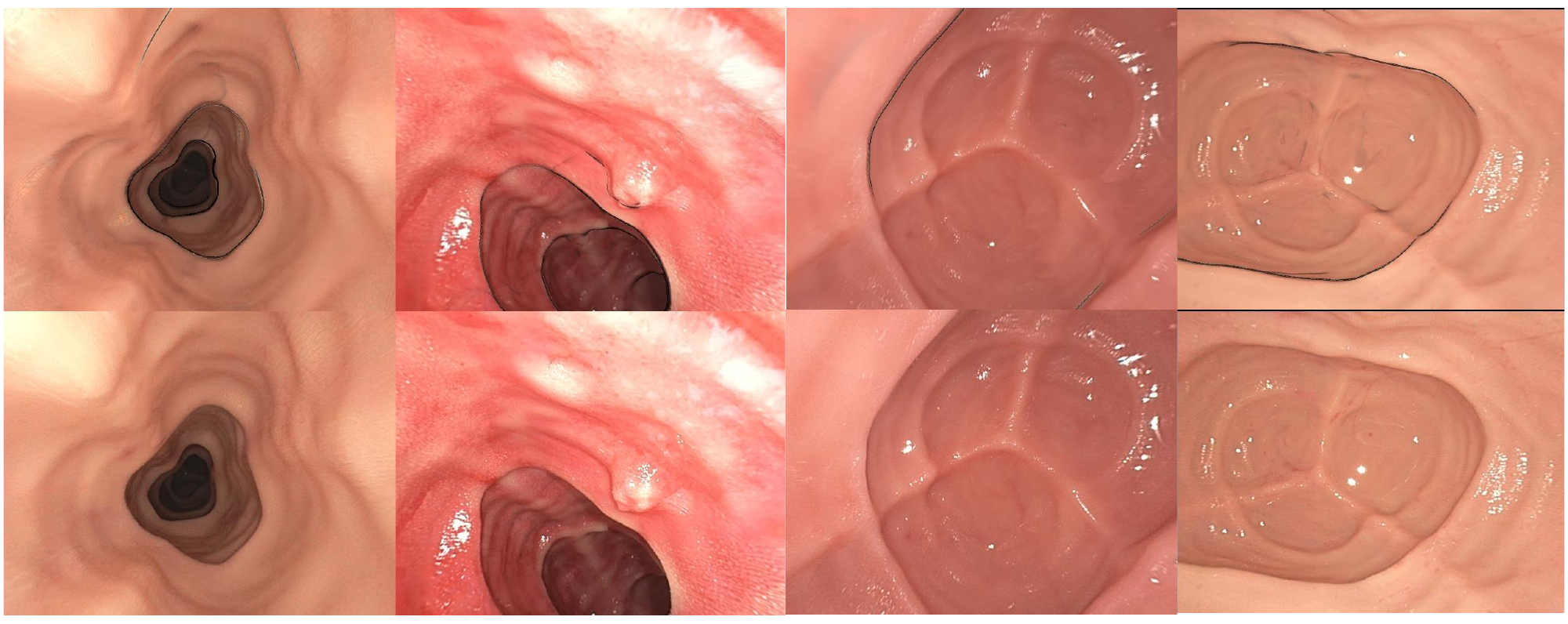}
    \caption{\textit{EndoPBR} generates photorealistic renderings from posed images and known geometry. Top row contains generated renderings from views outside the training set, and bottom row displays the ground truth RGB images. Note that we undistort images prior to the model training.}
    \label{fig:qual}
\end{figure}

Estimating geometry from RGB images, or vision-based 3D reconstruction, is a long-standing research problem and holds many applications in minimally invasive surgery (MIS). For instance, a Simultaneous Localization and Mapping (SLAM) system can lead to real-time surgical navigation guidance for endoscopy~\cite{liu2022sage}, enable robots to perceive the surgical scene~\cite{khan2024integration}, and actualize digital twins for various healthcare applications~\cite{ding2024digital}. However, surgical scenes are challenging to reconstruct with conventional algorithms due to factors like non-stationary lighting, textureless surfaces, and non-Lambertian surfaces. Most importantly, the lack of large labeled datasets for medical data limits 1) the utility of supervised learning of neural networks, and 2) a reliable platform to evaluate 3D reconstruction algorithms. Thus, existing research for vision-based 3D reconstruction in medical images falls into two main categories: 1) synthesizing a labeled dataset with simulated images or phantoms~\cite{mahmood2018deep,armin2017learning,rau2023bimodal}, or 2) training a neural network with self-supervision, e.g. joint ego-motion and depth estimation via photometric warping~\cite{liu2019dense,shao2022self,nazifi2024self}. Despite their impressive results, these methods have limited generalizability to patients and anatomies outside of the training set due to the inherent domain gap; synthetic data has limited photorealism and self-supervised models overfit to the training data. Recently, foundation models have emerged as potential solutions for medical vision tasks~\cite{rabbani2024can}, but their zero-shot performance in surgical scenes has room for improvement~\cite{han2024depth}. 

One promising avenue toward robust solutions for surgical 3D vision is synthetic labeled datasets of photorealistic images. If such a dataset could be generated at large-scale, then researchers could train models in a supervised manner or fine-tune foundation models for their specific anatomical domain. For example, synthetic data has been used for various medical computer vision tasks such as segmentation~\cite{rivoir2024importance,sahu2021simulation,venkatesh2024synthesizing}, depth estimation~\cite{mahmood2018deep,jeong2024depth}, and camera pose estimation~\cite{armin2017learning}. However, approaches that generate image frames independently are not suitable for systems like SLAM due to the lack of temporal and 3D consistency. Although some methods enforce such consistency with photometric warping~\cite{rivoir2021long} or learning mesh textures~\cite{han2024meshbrush}, these methods cannot generalize to novel views or lack photorealism. Phantom-based datasets like EndoSLAM~\cite{ozyoruk2021endoslam} and C3VD~\cite{bobrow2023colonoscopy} have been useful for the community, but they are not large enough to reliably train neural networks and limited to a specific anatomical domain. Recent works to generate realistic data use state-of-the-art neural reconstruction methods like Neural Radiance Fields (NeRFs)~\cite{mildenhall2021nerf} and 3D Gaussian Splatting~\cite{kerbl20233d} to perform novel view synthesis. However, these algorithms typically require comprehensive coverage of the scene at many camera vantage points, which is not typical in endoscopy~\cite{psychogyios2023realistic,bonilla2024gaussian}. Furthermore, vanilla versions of these algorithms assume stationary lighting and can be unstable with sparse views. To address these issues, we propose \textit{EndoPBR}, a photorealistic simulation platform for endoscopy that robustly generalizes to novel views. 

\textit{EndoPBR} generates photorealistic images by optimizing material properties and lighting conditions from images of known camera poses and geometry. To disambiguate the training process, we formulate domain-specific constraints of the scene. For material properties, we use the simplified Disney bidirectional reflective distribution function (BRDF) model~\cite{burley2012physically}, consisting of base albedo, metallic, and roughness for a given surface point. We parameterize the BRDF with a Multi-Layer Perceptron (MLP) neural network, which takes as input a query 3D point and predicts a 5D vector of material properties. For light conditions, previous works use a neural network to learn the surface light field with the given viewing direction and 3D point~\cite{yao2022neilf}. However, this formulation assumes stationary lighting and cannot account for varying illumination present in endoscopic scenes. To address this issue, \textit{EndoPBR} uses a moving spotlight model with learnable parameters to calculate incident light intensity at the surface. This formulation is simple enough for stable training but also allows for flexibility with learnable parameters to model the complexities of surgical scenes. We generate photorealistic outputs with physically-based rendering techniques and leverage differentiable rendering to minimize the difference between predicted and ground truth images.
\begin{figure}[htb]
\centering
    \includegraphics[width=0.9\linewidth]{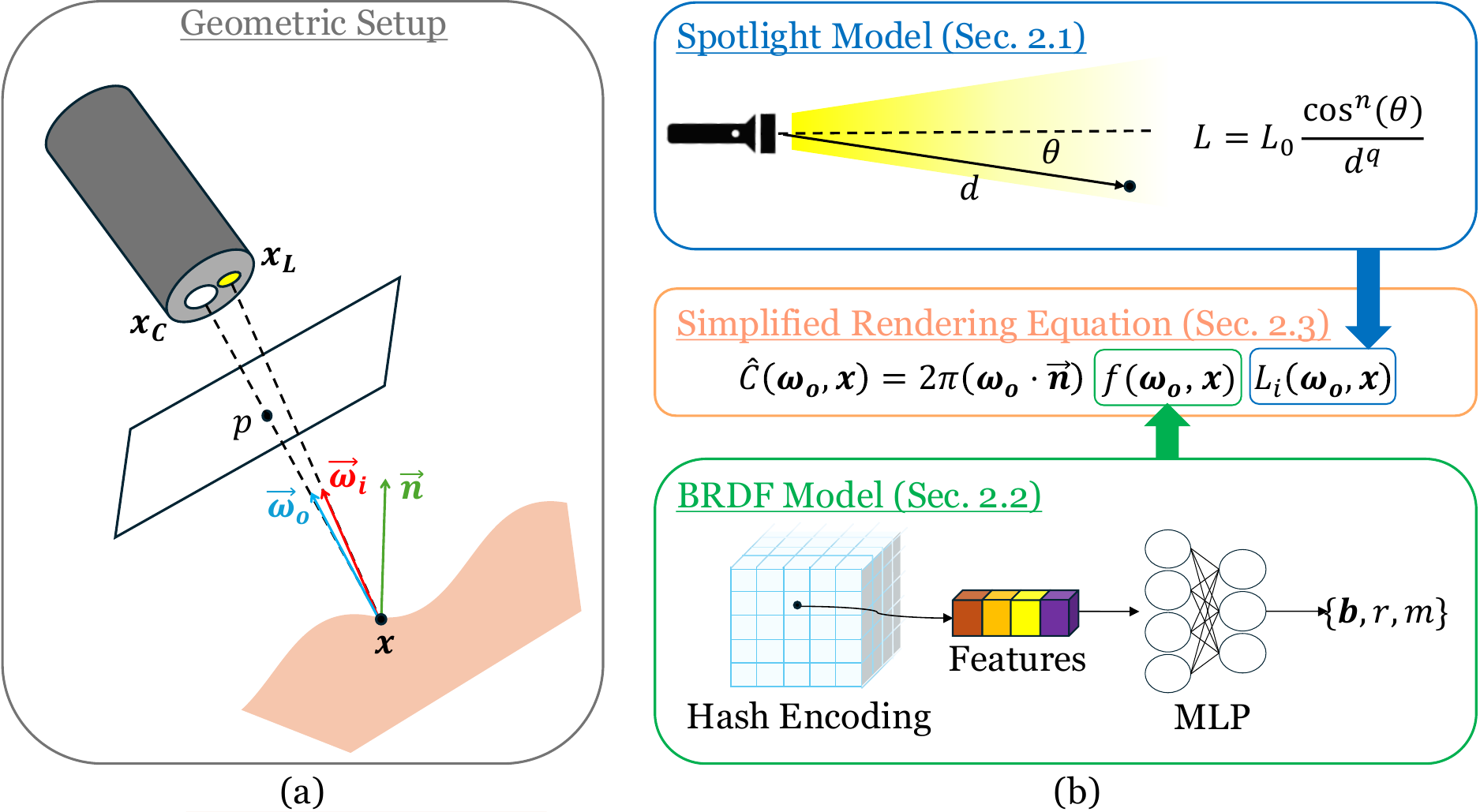}
    \caption{A description of the components of our pipeline. (a) describes the mathematical notation for the geometry of our setup. Given the camera center $x_c$, light source center $x_L$, and a query pixel $p_i$, we calculate its corresponding 3D point $\boldsymbol{x}$, which has an associated surface normal $\vec{\boldsymbol{n}}$, outgoing vector $\boldsymbol{\omega_o}$ and light incoming vector $\boldsymbol{\omega_i}$. (b) displays the essential components of our network. The learnable spotlight model is used to calculate the incident light intensity at $\boldsymbol{x}$ (Sec.~\ref{sec:light}), the BRDF model predicts material properties for $\boldsymbol{x}$ (Sec.~\ref{sec:brdf}), and these estimations are combined to predict the final pixel value via the rendering equation (Sec.~\ref{sec:simplification}).}
    \label{fig:workflow}
\end{figure}
We evaluate our algorithm on the Colonoscopy 3D Video Dataset (C3VD)~\cite{bobrow2023colonoscopy} on novel view synthesis by comparing generated renders with ground truth images. We show that by estimating material and incident light, we synthesize novel views with comparable performance with previous methods that use NeRF and 3DGS. Furthermore, we demonstrate the utility of photorealistic synthetic data by fine-tuning a depth estimation model with \textit{EndoPBR}'s generated images. We show that realistic synthetic data can be used to fine-tune a foundation depth estimation model, and that it produces similar results when fine-tuning with real images.

\section{Methods}
Our aim is to learn the material properties and lighting conditions of the scene given posed images and surface geometry, i.e. depth maps. An overview of our pipeline is shown in Fig.~\ref{fig:workflow}. \textit{EndoPBR} determines color pixel values with the physics of light transport via physically-based rendering. We demonstrate that our method generalizes to camera views outside the training set.

\textbf{Preliminaries}. The Rendering Equation~\cite{kajiya1986rendering} computes the outgoing radiance from a surface point $\boldsymbol{x}$ along a camera viewing direction $\boldsymbol{\omega_o}$:

\begin{equation}\label{eqn:rendering}
L_o(\boldsymbol{\omega_o}, \boldsymbol{x})=\int_\Omega f(\boldsymbol{\omega_o},\boldsymbol{\omega_i},\boldsymbol{x})L_i(\boldsymbol{\omega_i}, \boldsymbol{x})(\boldsymbol{\omega_i} \cdot \boldsymbol{n})d\boldsymbol{\omega_i}
\end{equation}

where $\boldsymbol{n}$ is the surface normal calculated from depth, $L_i$ is the incident light intensity coming from direction $\boldsymbol{\omega_i}$, and $f$ is the BRDF. The integration is performed in all incident directions $\boldsymbol{\omega_i}$ in the upper hemisphere $\Omega$ where $\boldsymbol{\omega_i} \cdot \boldsymbol{n} >0$. 

Given a pixel, we calculate its corresponding 3D point by unprojecting its depth value. We query its incident light intensity (Sec.~\ref{sec:light}), BRDF parameters (Sec.~\ref{sec:brdf}), and evaluate Eqn.~\ref{eqn:rendering} to estimate the pixel value (Sec.~\ref{sec:simplification}). We use differentiable rendering to learn the appropriate BRDF and scene lighting $L_i$ from endoscopic images.

\subsection{Light Conditions}\label{sec:light}
Decomposing input images to material and light properties is often difficult due to the training ambiguity. For instance, the network may model specularities (bright white spots on the surface) as white albedo instead of a reflective surface. To this end, properly constraining the model to learn valid light conditions is paramount for robust material estimation. For this purpose, we formulate key observations about surgical scenes as model learning constraints, specifically that 
\begin{enumerate}
    \item The light source moves in tandem with the camera, and
    \item there is no ambient light in surgical scenes; all visible light radiates from the endoscopic light source.
\end{enumerate}
Consequently, we model the environment lighting as a spotlight model located at $\boldsymbol{x}_L\in\mathbb{R}^3$ with direction $\boldsymbol{\vec{v}}_L \in \mathbb{R}^3$. Given a point $\boldsymbol{x}\in\mathbb{R}^3$, the incident light intensity $L_i$ experienced by $\boldsymbol{x}$ is
\begin{equation}\label{eqn:spotlight}
    L_i(\boldsymbol{x}) =L_0\frac{\cos^n(\theta)}{d^q}
\end{equation}
where $L_0$ is the spotlight base intensity, $\theta$ is the angle between $\boldsymbol{x}-\boldsymbol{x}_L$ and $\boldsymbol{\vec{v}}_L$, and $d$ is the distance from the light source to the point. We let $L_0$, $n$ and $p$ be learnable parameters to allow the model to flexibly model the environment lighting. Following previous work, we assume that the light source and camera center are co-located, i.e. $\boldsymbol{x}_c = \boldsymbol{x}_L$~\cite{batlle2023lightneus,psychogyios2023realistic}. We also set the light direction $\boldsymbol{v}_L$ equal to the camera forward viewing direction. 

\subsection{BRDF Estimation}\label{sec:brdf}
The BRDF is a fundamental concept in rendering and captures the interaction of light and matter. Following the Disney BRDF model~\cite{burley2012physically}, $f$ can be parameterized by a base color (or albedo) $\boldsymbol{b}\in [0,1]^3$, roughness $r\in[0,1]$, and metallic $m\in[0,1]$. We predict a 5D vector of these material properties for each spatial point by using a neural network, i.e. an MLP. We encode the spatial point with multiresolution hash encoding~\cite{muller2022instant} prior to the MLP to improve computational efficiency. 

We omit $\boldsymbol{x}$ in the following notation for simplicity. The BRDF can be decomposed into its diffuse and specular counterparts, $f = f_s+f_d$. The diffuse term equation is $f_d=\frac{1-m}{\pi}\boldsymbol{b}$ and the specular term is defined as

\begin{equation}\label{eqn:brdf_specular}
    f_s(\boldsymbol{\omega_i}, \boldsymbol{\omega_o})=\frac{D(\boldsymbol{h};r)F(\boldsymbol{\omega_i},\boldsymbol{h};\boldsymbol{b}, m)G(\boldsymbol{\omega_i},\boldsymbol{\omega_o},\boldsymbol{h};r)}{(\boldsymbol{n}\cdot \boldsymbol{\omega_i})(\boldsymbol{n}\cdot \boldsymbol{\omega_o})}
\end{equation}

where $\boldsymbol{h}$ is the half vector between $\boldsymbol{\omega_i}$ and $\boldsymbol{\omega_o}$, $D$ is the microfacets term, $F$ is the Fresnel term, and $G$ represents the GGX distribution. We use approximations of $D$, $F$, and $G$ used in previous work~\cite{yao2022neilf}, which we omit for the sake of brevity. 

\subsection{Domain-specific Simplifications}\label{sec:simplification} The rendering equation is typically computed by integrating over all incident light directions to account for global illumination. Although this costly summation is necessary in natural scenes, we assume that the majority of incident light comes from the spotlight source. Consequently, we omit the integral and compute pixel values only where $\boldsymbol{\omega_i}=\boldsymbol{\omega_o}$. As a result, we simplify Eqn.~\ref{eqn:rendering} and calculate the predicted color intensity as 

\begin{equation}\label{eqn:rendering_simplified}
\hat{C}(\boldsymbol{\omega_o}, \boldsymbol{x})=2\pi f(\boldsymbol{\omega_o},\boldsymbol{x})L_i(\boldsymbol{\omega_o}, \boldsymbol{x})(\boldsymbol{\omega_o} \cdot \boldsymbol{n})
\end{equation}

where $2\pi$ accounts for uniform Fibonacci sphere sampling from previous wokr~\cite{yao2022neilf}. We perform this calculation for each color channel to estimate RGB values. Finally, we constrain $f_s$ to produce white light, since the view-dependent colors in endoscopy are typically specularities. 

\subsection{Data Preparation and Loss Function}\label{sec:training}
In each iteration, we calculate a pixel's 3D point by unprojecting depth values with the camera intrinsics $\{f_x, f_y, c_x, c_y\}$ and extrinsics $\{R,t\}$. Specifically, given a pixel $(i,j)$ and its depth value $z$, we unproject that pixel as $x_c=(\frac{i-c_x}{f_x}z, \frac{j-c_y}{f_y}z, z)$ and transform it to world coordinates as $x_w=Rx_c+t$. We calculate surface normals by taking the image gradient of depth maps and using the camera extrinsics to project them into world space. Each point's material properties are estimated with the MLP and its incident light intensity from Eqn.~\ref{eqn:spotlight}, which are used to evaluate the rendering equation from Eqn.~\ref{eqn:rendering_simplified}. Finally, we apply a learned gamma correction to our rendered outputs to account for low dynamic range images common in endoscopic scenes, i.e. $\hat{C}^{LDR}=(\hat{C}^{HDR})^{\gamma}$.

\textbf{Loss}. We rely on the photometric L1 loss as the main objective to train \textit{EndoPBR}. Furthermore, we also impose material constraints, specifically that the metallic value $m(\boldsymbol{x})$ of all points should be minimized in endoscopic scenes and that the base color should not have abrupt changes. Our loss equation is 

$$\mathcal{L}=\mathcal{L}_1(\hat{C}, C) + \lambda_m |m(\boldsymbol{x})| + \lambda_b|\nabla_x b(\boldsymbol{x})|$$

\section{Experiments}\label{experiments}
We develop and test our system on C3VD, a colonoscopy dataset with ground truth poses and depth maps. We use their provided camera intrinsics to undistort the images and also resize them to $(480, 640)$. To ensure valid multiresolution hash encodings, we normalize all 3D points to $[0,1]$. 

\textbf{Implementation Details.} We sample $30$k pixels with a batch size of $5$ in each iteration. We train our model for $1500$ epochs using PyTorch on an NVIDIA RTX4090 GPU. The BRDF network's hidden dimensions are 64 with 2 layers following the multiresolution hash encoding layer~\cite{muller2022instant}. We use the Adam optimizer with learning rate $10^{-4}$ and $\beta=(0.9, 0.999)$. We set $\lambda_m = 10^{-4}$ and $\lambda_b=10^{-3}$ in our experiments. Finally, we apply a $8:1$ train-test split to evaluate our method, following previous work~\cite{bonilla2024gaussian}. Qualitative results of our method can be seen in Fig.~\ref{fig:qual}. 

We design two core experiments to demonstrate the photorealism and robustness of our model. First, we evaluate \textit{EndoPBR}'s ability to generalize to novel views with image quality metrics like Peak Signal-to-Noise Ratio (PSNR), Learned Perceptual Image Patch Similarity (LPIPS), and Structural Similarity (SSIM). We compare our results with vanilla NeRF~\cite{mildenhall2021nerf} as well as two recent works, namely REIM-NeRF~\cite{psychogyios2023realistic} and Gaussian Pancakes~\cite{bonilla2024gaussian} on the \verb|cecum_t4_b|, \verb|desc_t4_a|, \verb|trans_t1_a| sequences from C3VD. These works leverage state of the art neural reconstruction algorithms to synthesize novel views. 

\begin{figure}
    \centering
    \includegraphics[width=0.9\linewidth]{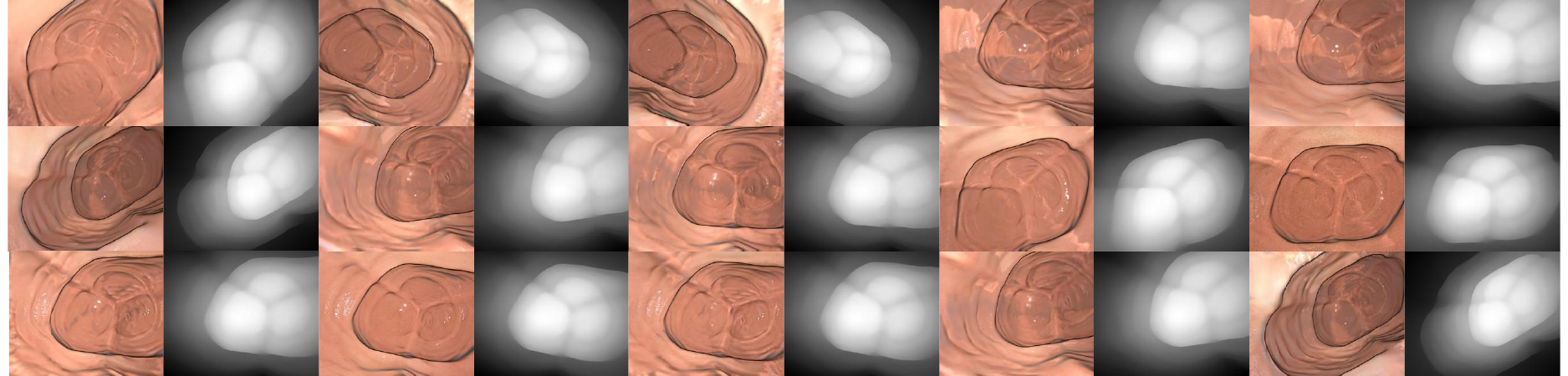}
    \caption{Examples of synthetic data produced by \textit{EndoPBR} to fine-tune Depth Anything V2. These images are generated by altering the camera view, material properties, or incident light intensity.}
    \label{fig:dataset}
\end{figure}

Second, we analyze how realistic synthetic data can be useful for developing 3D vision solutions in minimally invasive surgeries with a simple experiment: fine-tuning a foundation depth estimation model with rendered images. We first train \textit{EndoPBR} on the \verb|cecum_t1_b| sequence and apply specific transformations like camera translation and rotation, varied material properties, and adjusted light conditions. Because these components are explicitly disentangled in our model, we are able to adjust them separately for meaningful data augmentation compared with NeRF or 3DGS-based approaches. With the aforementioned transformations, we acquire a synthetic dataset of around $18$k image-depth-pose samples (from $765$ images), which we visualize in Fig.~\ref{fig:dataset}. We do not include any training camera views to ensure valid testing. 

We fine-tune a depth foundation model, namely ViT-S model of Depth Anything V2 (DAv2)~\cite{yang2025depth}, with the generated synthetic dataset for 100 epochs and the Adam optimizer (\verb|lr=5e-4|). The training set is completely comprised of rendered images, but we use original C3VD images to evaluate the depth predictions. We use conventional metrics like Absolute Relative Error, Root Mean Squared Error, and $\delta< 1.25$, the percentage of pixels within $25\%$ of ground truth depth values. We compare our results with baseline (zero-shot DAv2), baseline-scaled (median-scaled baseline), and using the original C3VD images to fine-tune the model. We note that original is the upper bound of depth estimation results. 

\section{Results and Discussion}

\textbf{Experiment 1. Novel View Synthesis.} Image quality metrics on the test set are shown in Table~\ref{tab:nvs}. The generated image quality results have PSNR and SSIM results on par with or better than REIM-NeRF and Gaussian Pancakes. However, our LPIPS score is notably worse than the other methods. We attribute this to some dark delineations present in rendered images, which can be seen in Fig.~\ref{fig:qual}. These black lines reflect the difficulty of resolving complex interactions between material properties and lighting in color calculation. However, our method is stable to train and does not rely on convergence of a different method as the input to our pipeline, as~\cite{bonilla2024gaussian} does.


\begin{table}
    \centering
    \caption{We compute the PSNR, LPIPS, and SSIM for generated images in the testing set for $3$ sequences and compare them with vanilla NeRF~\cite{mildenhall2021nerf}, REIM-NeRF~\cite{psychogyios2023realistic} and Gaussian Pancakes~\cite{bonilla2024gaussian}. We take numerical values directly from~\cite{bonilla2024gaussian}. }\label{tab:nvs}
    \begin{tabular}{l|c|c|c}
        \toprule
        Model & PSNR ($\uparrow$) & SSIM ($\uparrow$) & LPIPS ($\downarrow$)\\
        \hline 
        NeRF~\cite{mildenhall2021nerf} & 18.93 & 0.67 & 0.43 \\
        REIM-NeRF~\cite{psychogyios2023realistic} & 31.66 & 0.78 & 0.22 \\
        Gaussian Pancakes~\cite{bonilla2024gaussian} & 32.31 & 0.90 & 0.20 \\
        EndoPBR (Ours) & 30.39 & 0.86 & 0.30 \\
        \bottomrule
    \end{tabular}
\end{table}

\textbf{Experiment 2. Depth Estimation with Synthetic Data.} The values for depth estimation results are reported in Table~\ref{tab:depth}. First, we observe that the zero-shot depth estimation capabilities of DAv2 are imperfect even when median scaled to ground truth, which is supported by previous work~\cite{han2024depth}. Fine-tuning to a specific domain will be paramount to incorporate these foundation models to surgical 3D vision tasks. Furthermore, fine-tuning the model on synthetic data leads to similar results to fine-tuning with real images. This is demonstrated by the fact that all metrics are relatively similar between Original and our method. Because our synthetic data did not include any of the original training views or images, we claim leveraging simulated images is a promising venue to develop 3D vision tasks in endoscopy. To the best of our knowledge, this work is the first of its kind to evaluate the utility of synthetic data in this manner. 

\begin{table}
    \centering 
    \caption{We compare Abs. Rel., RMSE, and $\delta<1.25$ metrics between the baseline (zero-shot DAv2), baseline-scaled (zero-shot median scaled to ground truth used in previous work~\cite{han2024depth}, Original (fine-tuning with C3VD images), and our method (fine-tuning with \textit{EndoPBR}'s rendered images) on cecum t1 b.}   \label{tab:depth}
    \begin{tabular}{l|c|c|c}
        \toprule
        Method & Abs. Rel. ($\downarrow$) & RMSE ($\downarrow$) & $\delta <1.25$ ($\uparrow$)\\
        \hline 
        Baseline~\cite{yang2025depth} & $1.04\pm0.05$ & $0.22\pm0.00$ & $0.07\pm0.07$ \\
        Baseline Scaled~\cite{yang2025depth} & $0.31\pm 0.02$& $0.08\pm 0.01$& $0.45\pm0.06$\\
        Original Fine-Tune & $0.07\pm 0.01$ & $0.02\pm0.00$ & $1.00\pm0.00$ \\
        Ours Fine-Tune & $0.08\pm0.03$& $0.02\pm0.01$ & $0.95\pm0.09$ \\
        \bottomrule
    \end{tabular}
\end{table}

\textbf{Conclusion.} To summarize, we propose \textit{EndoPBR}, a endoscopic simulation platform that leverages differentiable rendering to estimate material properties and lighting conditions from posed images and geometry. We demonstrate that our model's novel view synthesis capabilities compete with existing methods that use NeRF and 3DGS while flexibly generalizing to novel views to synthesize a photorealistic dataset completely outside of its original training set. Furthermore, we show that realistic synthetic data generated from physics-informed learning is a promising avenue for robust 3D vision solutions in minimally invasive surgeries. We hope that this work inspires further exploration of generating realistic simulations. 

    

\begin{credits}
\subsubsection{\ackname} This study was funded by NIH T32 Training
Grant (grant number T32EB021937). 

\subsubsection{\discintname}
 The authors have no competing interests to declare that are
relevant to the content of this article. 
\end{credits}

%
%
%
\bibliography{mybib}
%




\end{document}